\documentclass{article}

\usepackage[final]{nips_2018}

\usepackage[utf8]{inputenc} 
\usepackage[T1]{fontenc}    
\usepackage{hyperref}       
\usepackage{url}            
\usepackage{booktabs}       
\usepackage{amsfonts}       
\usepackage{nicefrac}       
\usepackage{microtype}      
\usepackage{cleveref}
\usepackage{graphicx}		
\usepackage{algpseudocode} 
\usepackage{algorithm}
\algnewcommand\And{\textbf{ and }}
\usepackage{multicol}
\usepackage{natbib}

\title{Distributed traffic light control \\ at uncoupled intersections with real-world topology\\ by deep reinforcement learning}

\author{
	Mark Schutera \\
	Institute for Automation and Applied Informatics \\
	Karlsruhe Institute of Technology \\
	Research and Development \\
	ZF Friedrichshafen AG \\
	\texttt{mark.schutera@kit.edu} \\
	\And
	Niklas Goby\\
	Chair for Information Systems Research\\
	University of Freiburg \\
	IT Innovation Chapter Data Science \\
	ZF Friedrichshafen AG \\
	\texttt{niklas.goby@is.uni-freiburg.de} \\
	\And
	Stefan Smolarek\\
	IT Innovation Chapter Data Science \\
	ZF Friedrichshafen AG \\
	\texttt{stefan.smolarek@zf.com} \\
	\And
	Markus Reischl \\
	Institute for Automation and Applied Informatics \\
	Karlsruhe Institute of Technology \\
	\texttt{markus.reischl@kit.edu} \\
}

\begin{document}

\maketitle

\begin{abstract}
  This work examines the implications of uncoupled intersections with local real-world topology and sensor setup on traffic light control approaches. Control approaches are evaluated with respect to: Traffic flow, fuel consumption and noise emission at intersections. 
  The real-world road network of Friedrichshafen is depicted, preprocessed and the present traffic light controlled intersections are modeled with respect to state space and action space.  
  Different strategies, containing fixed-time, gap-based and time-based control approaches as well as our deep reinforcement learning based control approach, are implemented and assessed. Our novel DRL approach allows for modeling the TLC action space, with respect to phase selection as well as selection of transition timings. It was found that real-world topologies, and thus irregularly arranged intersections have an influence on the performance of traffic light control approaches. This is even to be observed within the same intersection types (n-arm, m-phases). Moreover we could show, that these influences can be efficiently dealt with by our deep reinforcement learning based control approach.
\end{abstract}

\section{Introduction}

Traffic, abstracted onto the macro level, underlies complex, time-varying \cite{TIMEDYNAMIC.1986} and stochastic \cite{STOCHASTICDYNAMIC.2002} dynamics. 
At the same time traffic management at intersections today is mostly conducted through rigid traffic regulations, such as right of way or pretimed traffic light signals. 
Ever increasing numbers of traffic participants result in a decreasing flow efficiency and other endemic traffic problems: Environmental pollution from the transport sector account to $15\%$ of overall greenhouse gas emissions \cite{EMISSION2010, EMISSION2014}. Long-term average levels of noise, in particular from traffic, cause harmful effects on human health \cite{ NOISEEUROPE.2002, NOISE2005}. Finally there is the problem of decreased traffic flow \cite{INTELLIGENTTRAFFIC.2001}. 
As of today, these issues are addressed by developing autonomous vehicles \cite{OVERVIEW.2017, Schutera.2018} or deploying intelligent infrastructures \cite{INTELLIGENTSYSTEM.2010,INTELLIGENTTRAFFIC.2001}, especially in the domain of traffic light control (TLC) \cite{Review2015}. 

\section{Related Work and Problem Statement}
\label{sec_sota}
As of today, local TLC, directed at uncoupled intersections, has been addressed by various approaches, which contribute to the progress in the area of dictating conflicting traffic movements efficiently. In the following strategies are distinguished between: Pretimed control, which is based on fixed control schemes that are predetermined offline, usually modeled by Webster's formula \cite{WEBSTER.1958}. Actuated control, in which based on traffic detectors a predefined control scheme is selected. Adaptive control, in which based on traffic detector data the priority is given to specific lanes (phase). Common schemes are MOVA \cite{MOVA.1988} and CRONOS \cite{CRONOS.2006}. 
However these approaches evince characteristic drawbacks \cite{REVIEWTLC.2018}. The pretimed controller lacks of adaptation with respect to changes in traffic volume, due to the fixed cycle-time. The actuated control overcomes this hindrance being able to switch between predefined control schemes, which are rarely ideal for the traffic flow demand at hand. In comparison, adaptive control strategies follow an optimization algorithm that determines the control action based on the entire intersection state. Hereby, adaptive strategies are prone to be incapable of real-time deployment.      
In current approaches \cite{SURVEYDRLTLC.2017}, distributed controllers are deployed at individual intersections, to increase real-time adaptation. Further, the state-action pairs are cast into Q-networks by data driven modeling through Deep Reinforcement Learning (DRL). TLC through DRL has proven to be operative and competitive on real-world data \cite{FLOW.2018, DRLTLC.2018,VANDERPOL.2016, INTELLILIGHT.2018}. 

State-of-the-art approaches neglect the topology of real-world intersections, addressing plain four-arm intersections with regular geometry, symmetric paths and uniform path lengths within the intersection and the road network. Further, the state space is often enhanced far above the actual detection capabilities of commonly deployed real-world sensor-setups, such as induction loop detectors (ILD). 
Increased traffic flow demand can be met by adapting the road network infrastructure itself (network reconstruction, road extension, roundabouts, etc.). However this is more than often not applicable at intersections with irregular topology, due to opposing boundary conditions (limited construction area). What is more, irregular topologies add to the complexity of the optimization problem of TLC schemes. Consequently, before anything else TLC needs to be approached with respect to irregular topologies under real-world conditions.  

The aim of this study is thus to examine the implications of real-world topologies and sensor-setups on the performance of DRL TLC approaches.
For this purpose the following stages have been undertaken:
\begin{itemize}
	\item Real-world intersection topologies 
	      are integrated into the micro-traffic simulation.
	\item A DRL model is deployed in order to cope with local real-world intersection topologies. 
	\item Ours and baseline TLC-approaches are evaluated with respect to traffic flow metrics. 
\end{itemize}

In Section \ref{sec:Problem}, the used framework and environment are outlined. Our method is presented in Section \ref{sec:Approaches} and the experimental results are depicted in Section \ref{sec:Experiments}. A conclusion is given in Section \ref{sec:Conclusion}.

\section{Framework}
\label{sec:Problem}

\subsection{Micro-Traffic Simulation of the Roadnetwork Friedrichshafen}
The road network is modeled as a bidirected graph with nodes (intersections) and edges (road segments).
For learning the TLC policy and training our DRL algorithm, we utilize the Simulation of Urban Mobility (SUMO) \cite{SUMO2012}.
SUMO is a free and open traffic simulation suite, which allows simulation of intermodal traffic (vehicles, pedestrians, etc.) and infrastructure (e.g. traffic lights). SUMO provides flexible APIs for road network design, traffic volume simulation and traffic light control.

\begin{figure}[ht]
	\centering
	\includegraphics[width=0.8\columnwidth]{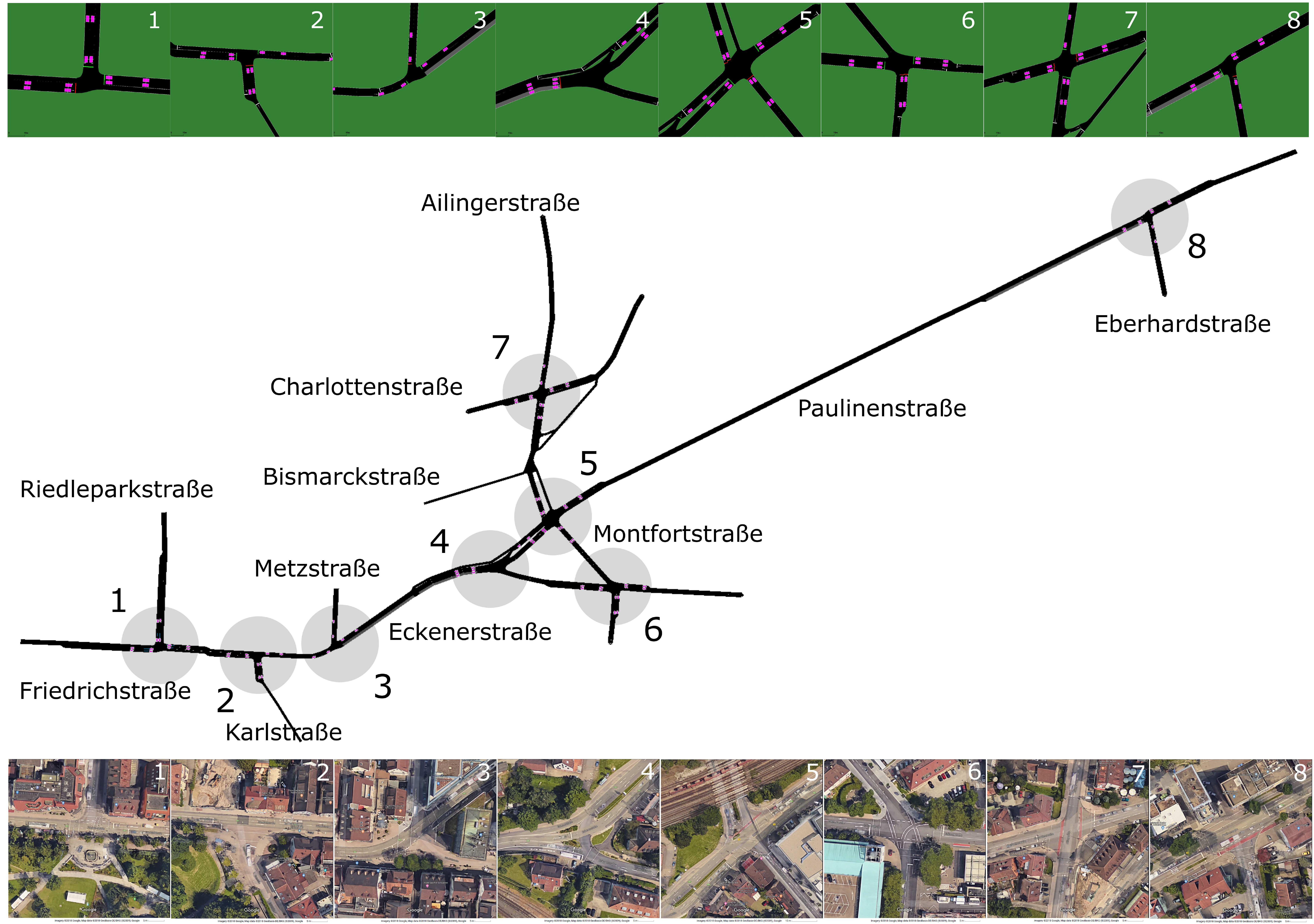}
	\caption{Overview of the Friedrichshafen roadnetwork with streetnames and the locations of the considered junctions in the center. In the top row the junctions are displayed as being present in SUMO. On the bottom row the google maps visualizations of the intersections themselves are shown.} 
	\label{fig:roadnetwork}
\end{figure}


The area of interest for the investigations is based on the real-world roadnetwork of the German city Friedrichshafen (see Fig.~\ref{fig:roadnetwork}). The sensor setup is based on ILDs, which are as of today the most commonly used means of measuring traffic in real-world applications, as is the case in Friedrichshafen. ILDs are installed in each lane of the intersection. An ILD samples, with frequency $f$, the number (flow) and duration (occupancy) of vehicles passing over \cite{ILD.2007, SUMO2012}.

The traffic flow has been generated following a probability distribution: In each simulation step a vehicle is instantiated with probability $0.2$ on a random lane. The vehicle are deployed with homogeneous characteristics: $maxVelocity=50~km/h$, $acceleration=3~m/s^2$, $deceleration=3~m/s^2$, $minGap=2.5~m$, $vehicleLength=5~m$.


\subsection{Traffic Light Control Approaches}

The basic unit of TLC strategies are non-conflicting vehicle movements linked together to phase groups \cite{GOV2009}. Vehicle movements are enabled or suppressed by traffic light signals ($Green$: Move with priority, $green$: Move without priority, $orange$: must yield, $red$: must stop) \cite{TLCBASICS.2012}. 
A phase group definition for $n$ traffic lights, thus consists of $n$ signals. As outlined in Sec. \ref{sec_sota}, there are a variation of TLC strategies. In the following three are considered as baseline for benchmarking, the implementation can be found within the TraCI python package\footnote{TraCI python package: \href{https://github.com/eclipse/sumo/tree/master/tools/traci}{https://github.com/eclipse/sumo/tree/master/tools/traci}}.

Fixed-time control generates a fixed cycle with a default cycle time (here: $31s$), for all traffic lights \cite{SUMO2012}.
All green phases are followed by an orange clearing phase of 6s. Left-turns are allowed (move without priority) at the same time as oncoming straight traffic.
	
Gap-based control is common in Germany and works through extending phase durations over $31s$ when a continuous stream of traffic is detected ($<3s$) or abbreviating phase durations when a sufficient time gap ($>3s$) in the stream is detected (within duration limits: $[6, 45]s$) \cite{SUMO2012}. This affects cycle duration in response to dynamic traffic conditions. This state space is determined by ILDs.

	
	
Time-based control, extends phase durations according to the presence of delayed vehicles \cite{SUMO2012}. For delay detection a detector that covers the full range of a lane is necessary (e.g. camera, extended ILDs). The delay of a vehicle is defined as $1 - \frac{v}{v_{max}}$, where v is its current velocity and $v_{max}$ the allowed maximal velocity on the lane. Once the $timeLoss$ oversteps the tolerated loss ($1s$), the corresponding $Green$ phase is extended by $timeToPass$, as such that the vehicle is able to pass the junction.
	
	
	
%

\section{Deep Reinforcement Learning for Traffic Light Control}
\label{sec:Approaches}

Deep reinforcement learning (DRL) combines the two frameworks deep learning (DL), for representation learning and reinforcement learning (RL), for learning to take actions in an initially unknown environment \cite{Silver.2016}. This new generation of algorithms has recently achieved human like results in mastering complex tasks in model-free settings with a very large state space and with no prior knowledge \citep{Mnih.2013,Mnih.2015,Silver.2017}.
For more background on RL, see \citep{Sutton.1998, Kaelbling.1996, Arulkumaran.2017}, for background on DL, see \citep{Goodfellow.2016c}, and for an overview of DRL, see \citep{Li.2017b}.

Following the general reinforcement learning setup, at each time-step $t$, the agent observes state $s$, extracted from the SUMO environment and takes action $a$ (i.e. the next phase configuration) according to a learned $\epsilon$-greedy strategy and receives reward $r$. The goal of the learning agent is to find a policy that maximizes the cumulative reward over: $R_t = \sum_{j=0}^{\infty} \gamma^j r_{t+j+1}$.

\paragraph{State Space}
Due to the irregular intersection geometry, the state space of the DRL problem needs to be variable in order to tackle the irregularity. For each intersection, we determine a set of all admissible phase configurations $P$, and a set $L$ of all induction loop detectors of all lanes. Each phase $p_k \in P$ can be uniquely assigned via an index $k$. 
Given these two sets, the state space $S$ for each intersection geometry is defined as tuple $s \in S $ and $s=(k, l_1, \dots, l_{|L|})$ where $k$ represents the 
index of the  intersections current traffic light phase $p_k$ , and $l_i$ the mean occupancy of the last time step in percent of the $i$th induction loop detector. The resulting state provided for the agent is a stack consisting of the last $\tau=10$ observed states $s_{t-\tau+1}, ..., s$.

\paragraph{Action Space}
The action space $A$ consists of integer values from the interval $\lbrack 0, \dots, |P|-1 \rbrack$. An action $a \in A$ corresponds to the phase transition from the current phase to phase $p_a$. To overcome crashes and emergency breaks of vehicles, we introduce an orange-phase for each lane to transition from green to red signals. This is done automatically and cannot be influenced by the agent. 

\paragraph{Reward Function}
The reward function plays an important role in learning the agent's behavior. Different reward functions lead to different, mostly unwanted, policies. Focusing solely on the delay or the waiting time as proxy for the average travel time of vehicles results in either a mixture of jammed and high speed lanes or flickering traffic lights, as shown by \cite{VANDERPOL.2016}. Therefore, they came up with a combination of these measures and included penalties for teleportation, emergency stops, and indicators for configuration changes \cite{VANDERPOL.2016}. A similar reward function is also used in \cite{INTELLILIGHT.2018}. Here, the authors added penalties for the queue length, the total number of vehicles and the travel time of vehicles that passed the intersection. Motivated by promising results of these two works, our one-step reward $r$ is calculated  by the sum of the average delay $d$ per lane, defined as $1 - \frac{v}{v_{max}}$, the average waiting time per lane $w$, a teleportation penalty $e$ and a flickering penalty $f$. The final one-step reward signal $r_t \in [-1 ;0]$ at time step $t$ is given by iterating over all lanes of the map and computing penalties, where $i$ is the lane index:  
\begin{equation}
r_t = -0.1f - 0.1e - 0.4\sum_{i=1}^{L}{d_i} - 0.4\sum_{i=1}^{L}{w_i},
\label{eq:reward}
\end{equation} 
where $L$ is the number lanes in the network. The coefficients of the reward (see Eq.~\ref{eq:reward}) are based on coefficients, which have been successfully deployed in similar studies \cite{VANDERPOL.2016, INTELLILIGHT.2018}.

\paragraph{Learning strategy}
The agent uses the Rainbow algorithm introduced in \cite{RAINBOW.2017}. Rainbow is based on the DQN algorithm \cite{Mnih.2013,Mnih.2015} which has been supplemented with additional components such as n-step Bellman updates \cite{NSTEP.2016}, prioritized experience replay \cite{EXPREPLAY.2015} and distribution reinforcement learning \cite{DISTRRL.2017}.


\section{Experiments}
\label{sec:Experiments}

\subsection{Implementation and Evaluation}

The experiments are developed in python using Google Dopamine \cite{DOPAMINE.2018} for implementing the agent and running the experiments and OpenAI gym \cite{OPENAIGYM.2016} as framework for creating the simulation environment that interacts with the agent. Each model has been trained on a NVIDIA K80 GPU or a NVIDIA M60 GPU, for approximately ten hours. 


Hyperparameters for the training process and the architecture have been based on previous studies, such as: The rainbow DRL approach \cite{RAINBOW.2017}, the C51 hyperparameter configuration \cite{DISTRRL.2017} and default values introduced by the dopamine framework \cite{DOPAMINE.2018}. The Q-network architecture is motivated by the DQN-architecture presented in \cite{Mnih.2015} and adapted to the present environment. 

\begin{table}
	\caption{Hyperparameters of the architecture and the training process of our DRL TLC approach. Hyperparameters not explicitly mentioned are set according to dopamine \cite{DOPAMINE.2018} or the C51 configuration presented in \cite{DISTRRL.2017}.}
	\label{sample-table}
	\centering
	\begin{tabular}{lll}
		\toprule
		\multicolumn{2}{c}{Architecture Q-Network}                   \\
		\cmidrule(r){1-2}
		Name    &  Value & Description \\
		\midrule
		Input Layer  & $len(s) \cdot \tau$  & Flattened    \\
		1. Hidden Layer    &  $2\cdot len(s)\cdot\tau$      & Fully connected, ReLU  \\
		2. Hidden Layer     &  $2\cdot len(s) \cdot\tau$     & Fully connected, ReLU  \\
		Output Layer   & $|A|$ & Fully connected  \\
		\midrule
		\multicolumn{2}{c}{Training}                   \\
		\cmidrule(r){1-2}
		Name        & Value \\
		\midrule
		Epsilon (Greedy policy) $\epsilon$  & $0.05$   \\
		Initial learning rate &   $0.0000625$     \\
		Epsilon (Adam optimizer)     & $0.00015$  \\
		Number of iterations            & $100$  \\
		Training steps   &        $36000$  \\
		Max. steps per episode          & $3600$  \\
		\bottomrule
	\end{tabular}
\end{table}

The baseline approaches, as well as our approach, are evaluated and compared with respect to minimizing $timeLoss$. It is assumed, that every vehicle desires to reach its destination with the least time expenditure possible. The $timeLoss$ is defined as the time expended due to driving below the $maxSpeed$. Thus, slowdowns due to stops at intersections will incur $timeLoss$ \cite{SUMO2012}.

\subsection{Results and Discussion}

\begin{figure}[ht]
	\centering
	\includegraphics[width=1\columnwidth]{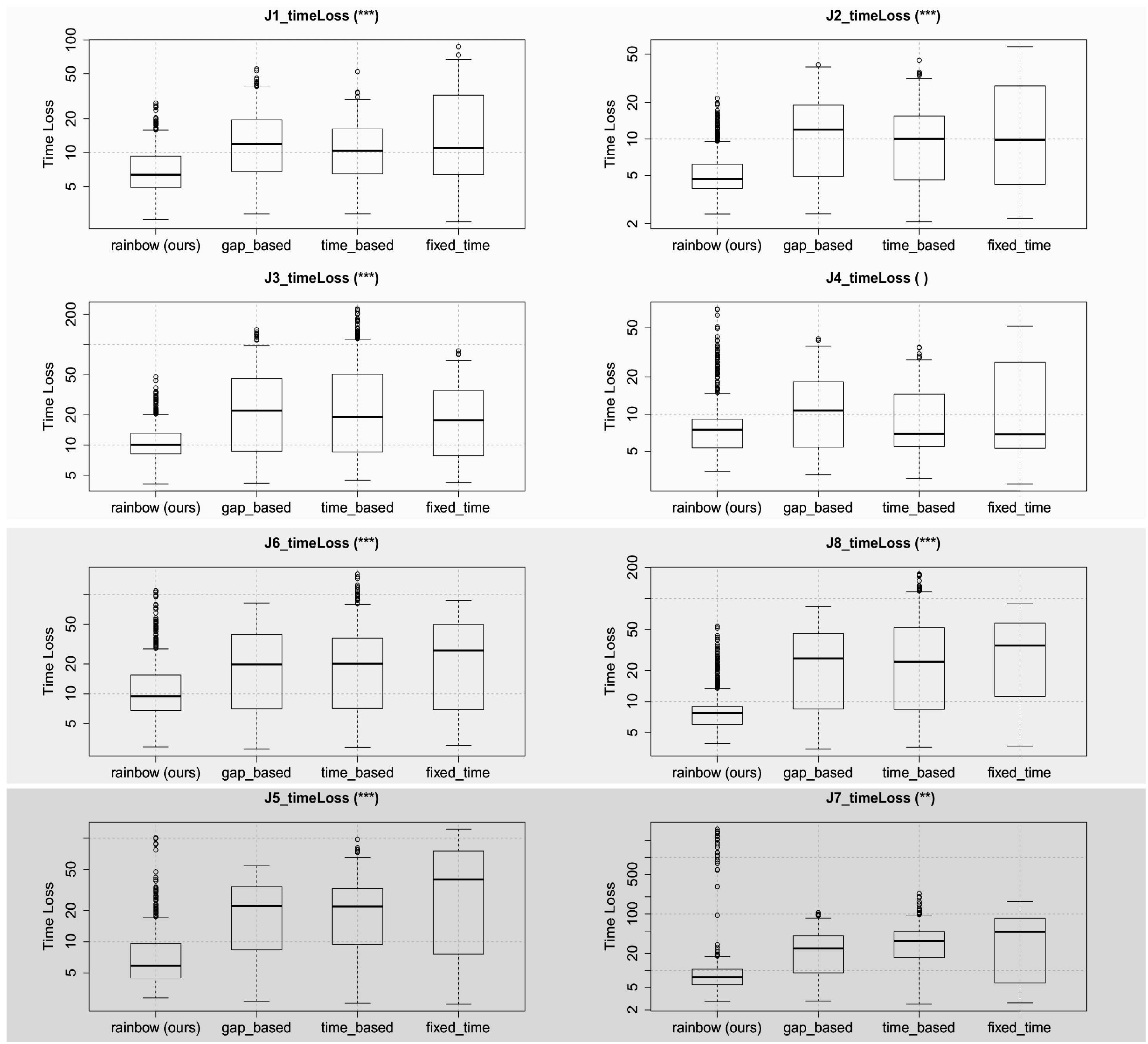}
	\caption{Evaluation with respect to the $timeLoss$ the vehicles experience due to the traffic light control at the junction (in seconds on a log-scale). The different shades mirror the characteristics of the junction $J$ whereas the number stands for the junctions ID (see~\ref{fig:roadnetwork}), from top to bottom: three-arm junction with four phases, three-arm junction with six states and four-arm junctions with eight states. The results of our Welch two sample t-test indicate that there is a statistically significant difference between the mean $timeLoss$ for our approach and the baseline approaches at nearly all junctions, displayed for each intersection next to the figure title.} 
	\label{fig:timeLoss}
\end{figure}

The approaches have been evaluated within SUMO, by running them on a test-sequence of 3600 simulation steps respectively. By comparing our results with the presented baseline methods with respect to the $timeLoss$ (see Sec.~\ref{sec:Approaches}), it is observed that (see Fig.~\ref{fig:timeLoss}):

The investigated real-world topologies show a critical influence on the DRL TLC performances (visible in both ours and the baseline). This is for one expressed by the differences of the mean $timeLoss$, ranging from $4-11$ seconds on intersections with the same amount of phases and arms (e.g. J1, J2, J3, J4), yet differing topologies. It is to be remarked that as of today the study of real-world topologies, with respect to DRL TLC, has been neglected and simplified within state-of-the-art research.

The interquartile range of our approach is smaller than six seconds, depicting the ability of our approach to have a smoothing effect on the majority of vehicle traffic flows. With the exception of the junction J4, our method clearly outperforms the baseline approaches by a difference in mean $timeLoss$ of five seconds at least. It is to be remarked, that our approach suffers from outliers, implying that the traffic flow is optimized at the expense of single vehicles. Nevertheless, those outliers are mostly within the range of the results achieved by the baseline approaches. 


\section{Conclusion}
\label{sec:Conclusion}

The influence of real-world topologies at uncoupled urban intersections in the presence of distributed TLC, has been investigated with respect to $timeLoss$. It was observed that topologies have a crucial impact on DRL TLC performance, yet being neglected in state-of-the-art research. It was demonstrated, that distributed TLC strategies are applicable at uncoupled intersections with real-world topology. A beneficial effect of such strategies with respect to the $timeLoss$, has been shown compared to baseline algorithms. 
Also to be highlighted is the novel modeling of the DRL TLC action space, which allows for a phase selection rather than the mere selection of transition timings, latter being prevalent in the state-of-the-art. 
SUMO is a vast simplification of real-world traffic. Introducing pedestrians and cyclists to the environment, as well as emulating traffic flow with respect to real-world data, is thus considered the next step to get closer to real-world conditions. Future work should also investigate the influence of differing sensor setups and thus extensions of the observed state space of the DRL TLC. This might include adding information about daytime, date, current weather or road friction, up to an omniscient state perception.
It is to be remarked that there is no guarantee, that the sum of locally optimized strategies also leads to a global optimum when being coupled. Our ongoing work thus seeks to investigate the implications of coupled intersections and the interaction between distributed systems. A deeper understanding of underlying action patterns of the DRL TLC should contribute to further functional validation, that's why we suggest a future action space pattern analysis. Judging from our results, it can be noted that the reward function matters most. Further analysis aiming at reducing outliers is highly recommended.



\subsubsection*{Acknowledgments}

We thank Dr. Jochen Abhau from Research and Development, as well as the whole Data Science Team at ZF Friedrichshafen AG, for supporting this research. Thank you for all the assistance and comments that greatly improved this work. We would also like to express our gratitude to Prof. Dr. Ralf Mikut from the Institute for Automation and Applied Informatics, Karlsruhe Institute of Technology, and Prof. Dr. Dirk Neumann for providing insight and expertise that greatly enhanced our research. 

\bibliographystyle{plain}
\bibliography{references}

\end{document}